\documentclass{article}

\usepackage{amsmath,amsfonts,bm}

\def\eqref#1{equation~\ref{#1}}

\def\1{\bm{1}}

\def\va{{\bm{a}}}

\def\vg{{\bm{g}}}

\def\vs{{\bm{s}}}

\def\vv{{\bm{v}}}
\def\vw{{\bm{w}}}
\def\vx{{\bm{x}}}
\def\vy{{\bm{y}}}

\def\mA{{\bm{A}}}

\def\mE{{\bm{E}}}

\def\mG{{\bm{G}}}

\def\mT{{\bm{T}}}

\def\mW{{\bm{W}}}

\DeclareMathAlphabet{\mathsfit}{\encodingdefault}{\sfdefault}{m}{sl}
\SetMathAlphabet{\mathsfit}{bold}{\encodingdefault}{\sfdefault}{bx}{n}

\newcommand{\E}{\mathbb{E}}

\usepackage[accepted]{icml2026}
\usepackage{url}

\usepackage{times}
\usepackage{amsmath}
\usepackage{amsthm}
\usepackage{graphicx}
\usepackage{booktabs}
\usepackage{tabularray}
\usepackage{array}
\usepackage{ragged2e}
\usepackage{multirow}
\usepackage{amsfonts}
\usepackage{colortbl}
\usepackage{setspace}
\usepackage{caption} 
\usepackage{subcaption}
\usepackage{wasysym}
\usepackage{longtable}
\usepackage{array}
\usepackage{diagbox}  
\usepackage{caption}
\usepackage{comment}
\usepackage{float}
\usepackage{enumitem}

\usepackage{pifont}
\usepackage{booktabs}

\usepackage[capitalize,noabbrev]{cleveref}

\usepackage[textsize=tiny]{todonotes}

\icmltitlerunning{ExpertWeaver: Unlocking the Inherent MoE in Dense LLMs with GLU Activation Patterns}

\begin{document}

\twocolumn[
  \icmltitle{ExpertWeaver: Unlocking the Inherent MoE in Dense LLMs with GLU Activation Patterns}



  \icmlsetsymbol{equal}{*}

  \begin{icmlauthorlist}
    \icmlauthor{Ziyu Zhao}{ZJU,SII}
    \icmlauthor{Tong Zhu}{AILAB}
    \icmlauthor{Zhi Zhang}{Seed}
    \icmlauthor{Tiantian Fan}{Seed}
    \icmlauthor{Jinluan Yang}{ZJU}
    \icmlauthor{Kun Kuang}{ZJU}
    \icmlauthor{Zhongyu Wei}{FDU}
    \icmlauthor{Fei Wu}{ZJU}
    \icmlauthor{Yu Cheng}{CUHK}
  \end{icmlauthorlist}

  \icmlaffiliation{ZJU}{Zhejiang University}
  \icmlaffiliation{AILAB}{Shanghai AI Laboratory}
  \icmlaffiliation{SII}{Shanghai Innovation Institute}
  \icmlaffiliation{Seed}{Bytedance Seed}
  \icmlaffiliation{FDU}{Fudan University}
  \icmlaffiliation{CUHK}{Chinese University of Hong Kong}

  \icmlcorrespondingauthor{Ziyu Zhao}{benzhao.styx@gmail.com}

  \icmlkeywords{Machine Learning, ICML}

  \vskip 0.3in
]



\printAffiliationsAndNotice{}  

\begin{abstract}
Mixture-of-Experts (MoE) effectively scales model capacity while preserving computational efficiency through sparse expert activation.
However, training high-quality MoEs from scratch is prohibitively expensive.
A promising alternative is to convert pretrained dense models into sparse MoEs.
Existing dense-to-MoE methods fall into two categories: \textbf{dynamic structural pruning} that converts dense models into MoE architectures with moderate sparsity to balance performance and inference efficiency, and \textbf{downcycling} approaches that use pretrained dense models to initialize highly sparse MoE architectures. However, existing methods break the intrinsic activation patterns within dense models, leading to suboptimal expert construction.
In this work, we argue that the Gated Linear Unit (GLU) mechanism provides a natural blueprint for dense-to-MoE conversion. We show that the fine-grained neural-wise activation patterns of GLU reveal a coarse-grained structure, uncovering an inherent MoE architecture composed of consistently activated universal neurons and dynamically activated specialized neurons.
Leveraging this discovery, we introduce ExpertWeaver, a training-free framework that partitions neurons according to their activation patterns and constructs shared experts and specialized routed experts with layer-adaptive configurations.
Our experiments demonstrate that ExpertWeaver significantly outperforms existing methods, both as a training-free dynamic structural pruning technique and as a downcycling strategy for superior MoE initialization.
\end{abstract}

\section{Introduction}

The Mixture-of-Experts (MoE) architecture effectively expands model capacity while maintaining efficiency by activating only sparse subsets of experts. However, training MoE models from scratch is rather expensive, driving growing interest in converting pretrained dense models into sparse MoE architectures. These efforts can be broadly categorized into two distinct lines of research based on target sparsity and objectives.
The first one is \textit{Dynamic Structural Pruning}, which converts dense models into MoE architectures with moderate sparsity (e.g., 25\%) to balance performance and efficiency \citep{gao2025tomoe,pei2025cmoe,nishu2025dense,zheng2024learn}. The second category is \textit{Downcycling} (versus \textit{upcycling} that expands models \citep{he2024upcycling,komatsuzaki2022sparse}), which initializes highly sparse MoEs (75\%+ sparsity) by splitting pretrained dense MLPs into smaller experts, avoiding the prohibitive costs of training sparse MoEs from scratch \citep{zhang2022moefication,zhu2024llama,qu2024llama}.
However, existing methods break the inherent activation patterns, necessitating additional router training, and ignore layer-specific differences, leading to suboptimal expert construction.
The detailed related work is shown in Appendix~\ref{app:related_work}.

\begin{figure*}
    \centering
\includegraphics[width=\linewidth]{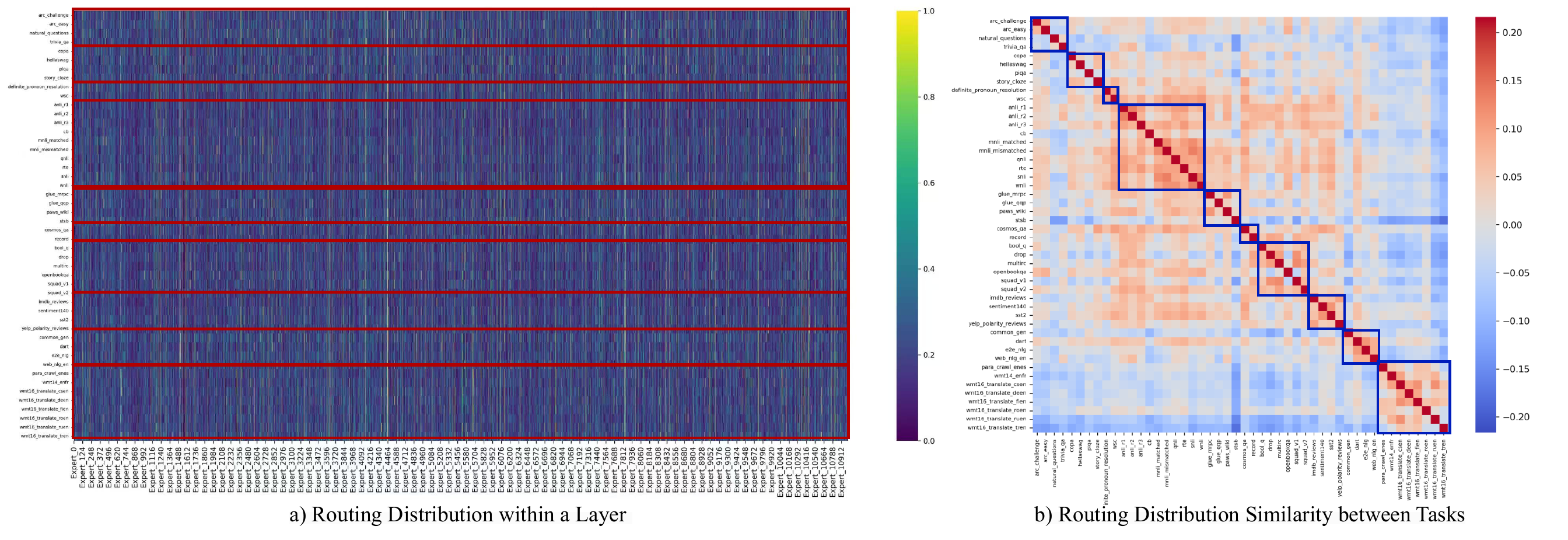}
    \caption{\textbf{Neuron activation patterns across diverse tasks.} We visualize the middle layer's activation patterns from Qwen2.5-7B on a subset of Flan-v2. a) Activation distribution of neurons across different tasks. b) Activation distribution of neurons within individual task clusters, where tasks belonging to the same cluster are enclosed in boxes.}
    \label{fig:motivation2}
\end{figure*}

In this work, we argue that the key to overcoming these limitations lies in the GLU mechanism \citep{shazeer2020glu}, which provides a natural blueprint for dense-to-MoE conversion by revealing the model's intrinsic functional structure.
The GLU mechanism employs additional gating weights to dynamically control neuron activations based on the input context, providing meaningful indicators of each neuron's role and importance for effective MoE expert construction.
Our analysis of the GLU mechanism in \S\ref{sec:motivation} yields key insights that serve as the cornerstone of our method. Firstly, GLU gating signals naturally induce token-level activation sparsity (\textbf{Takeaway 1}). Moreover, these patterns reveal both universal neurons ideal for a shared expert and task-specific clusters suitable for routed experts (\textbf{Takeaways 2 \& 3}). Additionally, by quantifying neuron specialization with the \textit{Coefficient of Variation (CV)}, we find that the distribution of universal versus specialized neurons varies systematically across layers, indicating that different layers require different expert configurations rather than a one-size-fits-all approach (\textbf{Takeaway 4}). Together, these observations provide a complete blueprint for constructing a coarse-grained, expert-level MoE architecture from fine-grained, neural-level GLU activations.


Based on these findings, we propose \textbf{ExpertWeaver}, a novel, training-free method that effectively converts a dense model into MoE guided by GLU activation patterns. The process consists of three key stages.
a) Firstly, the activation pattern of all neurons is captured by recording their GLU activations from a multi-task calibration dataset.
b) Then, the CVs of these activations are calculated to perform a layer-aware configuration, determining the precise size of the shared expert and the pool of routed experts for each layer.
c) Finally, neurons are partitioned according to layer-wise configuration by grouping the most consistently active neurons into a single \textit{shared expert}, while remaining specialized neurons are clustered into \textit{routed experts} based on their activation patterns. The MoE router is also constructed in a training-free manner from the GLU gating centroids.
This entire process successfully ``weaves'' the neurons of a dense model into multiple experts, constructing a well-structured MoE architecture.

We conduct extensive experiments to validate the effectiveness of ExpertWeaver across two categories of methods. As a training-free dynamic structural pruning method, ExpertWeaver significantly outperforms existing structural pruning baselines on multiple benchmarks. Furthermore, when applied as a downcycling strategy, our method successfully initializes a highly sparse MoE model that, with limited continued pretraining, achieves superior performance compared to both established MoE and dense baselines with comparable parameters and training budgets.
The main contributions of this paper are as follows:
\begin{itemize}
    [leftmargin=*,itemsep=1pt]
    \item We systematically analyze GLU activation patterns in dense LLMs and find that their fine-grained activation patterns exhibit structural properties that provide a natural blueprint for constructing MoEs with coarse-grained activation.
    
    \item We propose ExpertWeaver, a novel training-free framework that leverages GLU activation patterns to convert dense models into MoE architectures.
    
    \item Comprehensive experiments demonstrate that ExpertWeaver significantly outperforms existing methods in both structural pruning and MoE downcycling across diverse downstream tasks.
    \end{itemize}

\section{Preliminaries}

\subsection{Background}
\label{sec:preliminary}
\paragraph{Gated Linear Unit.}
Most current high-performance LLMs have adopted the GLU architecture, with SwiGLU being the most commonly used variant \citep{shazeer2020glu}. A dense GLU layer processes an input $x$ using three distinct weight matrices: $\mW_{\text{gate}}$, $\mW_{\text{up}}$, and $\mW_{\text{down}}$. The entire computation can be expressed as:
$$
\vy = \left( \text{Swish}(\vx \mW_{\text{gate}}) \odot (\vx \mW_{\text{up}}) \right) \mW_{\text{down}}.
$$
The element-wise product $\odot$ combines the input projection with a dynamic gate that modulates neuron activation strength for each token. This gating mechanism provides dense GLU models with intrinsic dynamic sparsity, which we identify as the key to efficiently converting pretrained dense models into structured sparse MoEs.

\paragraph{Mixture-of-Experts.}
The MoE architecture replaces dense Feed-Forward Networks (FFNs) in LLMs with multiple smaller expert networks and a gating mechanism for sparse activation. Specifically, an MoE layer consists of $N$ experts ($\mE_1, \dots, \mE_N$) and a router network $\mG$ that dynamically selects which experts to activate for each input token:
\begin{equation}
\begin{split}
    \vy &= \sum_{i \in \text{TopK}(\mG(\vx))} \vg(\vx)_i \mE_i(\vx); \\
    \mE_i(\vx) &= \left( \text{Swish}(\vx \mW_{\text{gate}}^{(i)}) \odot (\vx \mW_{\text{up}}^{(i)}) \right) \mW_{\text{down}}^{(i)},
\end{split}
\end{equation}
where $\vg(x)_i$ represents the normalized gating weight for expert $i$, and each expert has its own parameters. By activating only a subset of experts per token, MoE maintains model capacity while significantly reducing computational costs.

\subsection{Motivation}
\label{sec:motivation}
As shown in \S\ref{sec:preliminary}, GLU and MoE represent two distinct granularities of activation patterns. GLU implements fine-grained activation at the neuron level through a dynamic gate that modulates neuron activation strength for each token. In contrast, MoE employs coarse-grained activation by routing entire tokens to selected experts, establishing macroscopic, structural activation at the expert level. This difference between neural-level and expert-level activations raises an intriguing question: \textit{can the fine-grained neuron-level activation patterns of GLU be aggregated to inform the construction of a coarse-grained, expert-level activated MoE architecture?} Our exploratory analysis of the GLU activation patterns in pretrained dense LLMs yields four key insights that form the cornerstone of \textbf{ExpertWeaver}.

\subsubsection{GLU as a Natural Sparsity Signal}

As established in \S\ref{sec:preliminary}, the gating mechanism in a GLU, $\text{Swish}(\vx \mW_{\text{gate}})$, acts as a dynamic, data-dependent filter. To investigate the feasibility of leveraging inherent signals to directly enforce structured sparsity, we introduce AbsTopk-GLU, a simple modification that explicitly introduces sparsity into FFNs based on their gating activation scores:
\begin{equation}
\label{eq:topk_glu}
\text{AbsTopk-GLU}(\vx) = \left( \text{AbsTopK}(\text{Swish}(\vx \mW_{\text{gate}}), k) \right) \odot (\vx \mW_{\text{up}}).
\end{equation}
Here, $\text{AbsTopK}(\vv, k)$ preserves the top $k$ values in $|\vv|$ while zeroing out the rest, effectively retaining only the most activated neurons for each token.

\begin{table}
    \centering
    \caption{\textbf{Performance of AbsTopk-GLU vs. unstructured pruning at 50\% sparsity.} (Numbers in parentheses denote the number of shots for evaluation; no annotation means zero-shot.)}
    \label{tab:unstructured_pruning}
    \resizebox{\linewidth}{!}{%
\begin{tabular}{lcccccc}
    \toprule
    Method & MMLU(5) & HellaSwag(10) & ARC-e & ARC-C(25) & PiQA & Avg. \\
    \midrule
    \multicolumn{7}{c}{\small LLaMA3-8B, 50\% sparsity} \\
    \hline
    Dense       & 65.3 & 82.1 & 77.9 & 57.9 & 80.8 & 72.8 \\
    Wanda       & 55.8 & 75.0 & 72.0 & 51.3 & 77.3 & 66.3 \\
    SparseGPT   & 57.4 & 75.5 & 72.0 & 50.3 & 78.1 & 66.7 \\
    Magnitude   & 48.3 & 41.9 & 53.3 & 38.7 & 65.8 & 49.6 \\
    \rowcolor{gray!10}
    AbsTopk-GLU    & \textbf{58.8} & \textbf{79.4} & \textbf{72.4} & \textbf{51.6} & \textbf{78.6} & \textbf{68.2} \\
    \midrule
    \multicolumn{7}{c}{\small Qwen2.5-7B, 50\% sparsity} \\
    \hline
    Dense       & 74.2 & 80.3 & 77.8 & 63.8 & 80.0 & 75.2 \\
    Wanda       & 69.2 & 74.1 & 75.5 & 55.5 & 78.1 & 70.5 \\
    SparseGPT   & 69.8 & 75.9 & 74.7 & 56.7 & 78.3 & 71.1 \\
    Magnitude   & 64.2 & 49.5 & 46.3 & 33.6 & 63.8 & 51.5 \\
    \rowcolor{gray!10}
    AbsTopk-GLU    & \textbf{70.8} & \textbf{78.6} & \textbf{76.6} & \textbf{59.7} & \textbf{78.9} & \textbf{72.9} \\
    \bottomrule
\end{tabular}
    }
\end{table}
\textbf{Takeaway 1: The GLU's gating mechanism naturally induces structured sparsity within dense LLMs.}
As shown in Table~\ref{tab:unstructured_pruning}, \textit{AbsTopK-GLU}, which activates only 50\% of neurons, consistently and significantly surpasses established unstructured pruning methods such as Wanda \citep{sun2023simple} and SparseGPT \citep{frantar2023sparsegpt}. This performance highlights the inherent structural sparsity of dense models and validates the use of GLU gate scores as a reliable mechanism for controlling activation patterns.
However, this fine-grained expert partitioning, where each neuron is treated as an individual expert, is impractical for hardware implementation. While it reduces theoretical FLOPs, the overhead of TopK selection and scattered memory access prevents real-world applications. Nonetheless, this experiment reveals the crucial insight that the GLU gating signal's control over neuron-level activation can be extended to manage coarse-grained experts, motivating our approach of grouping neurons into larger, hardware-friendly blocks.


\subsubsection{Identifying Universal and Specialized Neurons via GLU Activation Patterns}
\label{sec:gating_patterns}
To investigate the structure of GLU activation patterns, we analyzed activation patterns recorded for 5 few-shot samples per task in the Flan-v2 collection (48 tasks, 10 task clusters; details in Appendix~\ref{app:flan_details}). The results in Figure~\ref{fig:motivation2} clearly illustrate two distinct and complementary phenomena, which inform the following observations.

\textbf{Takeaway 2: There exists a core set of universally important neurons that are consistently activated across different tasks.} As shown in Figure~\ref{fig:motivation2}a), we observed that a consistent subset of neurons showed high activation regardless of the task domain. We hypothesize that these neurons encode task-agnostic knowledge. This observation motivates the design of MoE architectures that incorporate a \textit{shared expert} specifically dedicated to capturing and leveraging this task-agnostic knowledge, thereby enhancing both parameter efficiency and overall performance.

\textbf{Takeaway 3: Specialized neurons exhibit task-specific co-activation patterns.} As illustrated in Figure~\ref{fig:motivation2}b), the activation pattern similarity heatmap, generated from truncated and normalized neuron activations, reveals clear block-diagonal structures, where semantically related tasks show high similarity in specialized neuron activation patterns. This demonstrates that neurons naturally form co-activation clusters organized by task semantics, providing a principled approach for constructing \textit{routed experts} through clustering based on multi-task activation patterns.

\textbf{Takeaways 2\&3} collectively provide a blueprint for dense-to-MoE conversion by forming shared experts from universal neurons and constructing routed experts through clustering specialized neurons based on their co-activation patterns.

\begin{figure}
\centering
\includegraphics[width=0.65\linewidth]{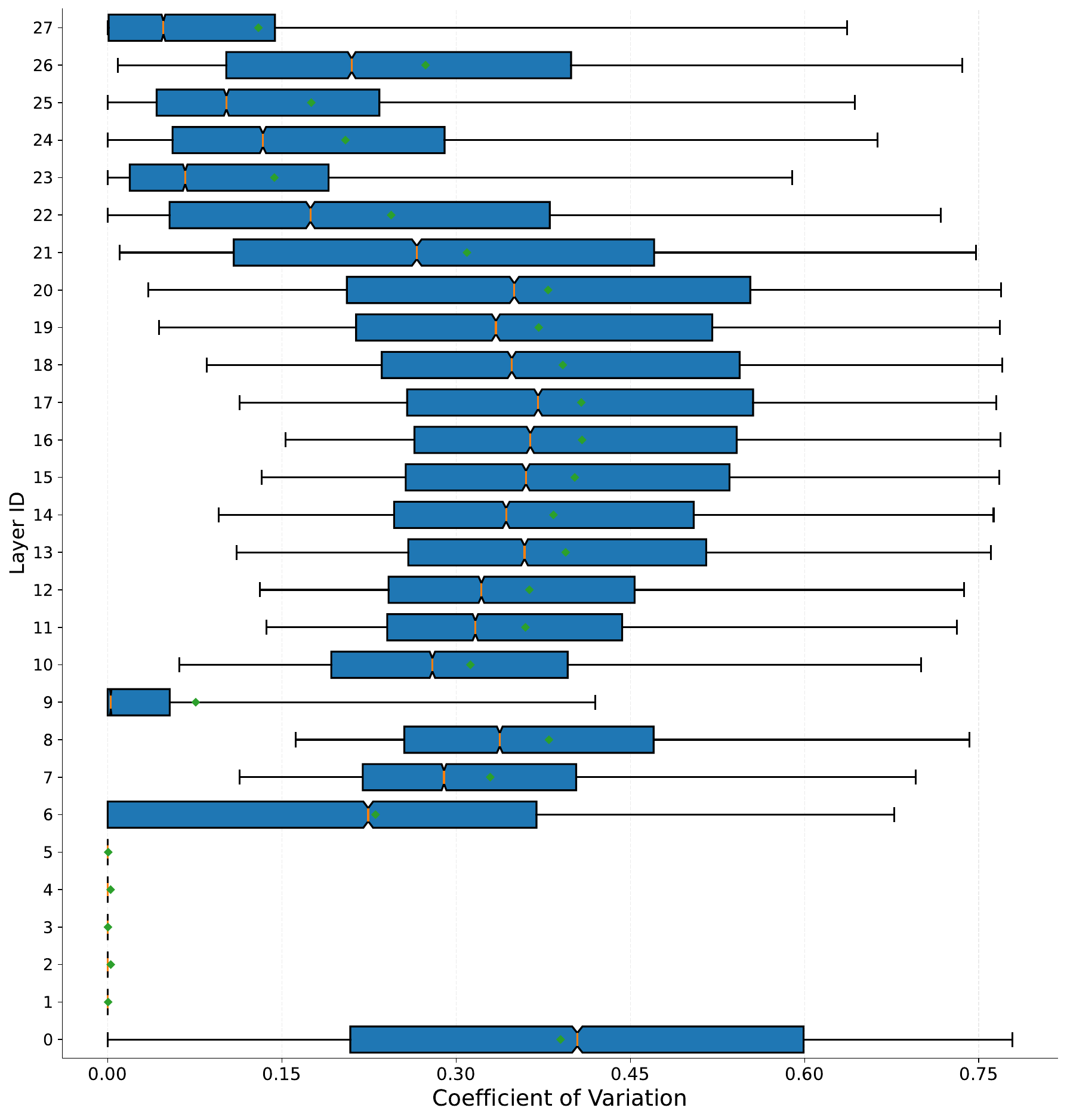}
\caption{\textbf{Neuron Coefficient of Variation Across Layers}.}
\label{fig:cv_plot_box}
\end{figure}

\subsubsection{Layer-Aware Expert Allocation}
Building on our discovery of universal and specialized neurons in \S\ref{sec:gating_patterns}, we next investigate how the ratio of universal to specialized neurons varies across layers.
To quantify this layer-wise behavior, we measure each neuron's activation consistency using the Coefficient of Variation (CV), which is defined as the ratio of the standard deviation to the mean \citep{abdi2010coefficient}:
\begin{equation}
    \label{eq:cv_def}
    \mathrm{CV}(a_j) = \frac{\sigma_j}{\mu_j+\epsilon},
\end{equation}
where $\mu_j = \E_{t \in \mathcal{T}}[\bar{a}_{j,t}]$ and $\sigma_j = \text{STD}_{t \in \mathcal{T}}[\bar{a}_{j,t}]$ are the mean and standard deviation of the neuron's average absolute activation, $\bar{a}_{j,t}$, across the tasks in our calibration set $\mathcal{D}_{\text{calib}}$. A low CV indicates a \textit{universal neuron} with consistent activation, while a high CV points to a \textit{specialized neuron} that activates selectively.


\textbf{Takeaway 4: Layers exhibit different levels of neuron specialization.} Figure~\ref{fig:cv_plot_box} shows that boundary layers (shallow and deep) consistently have low CV scores, indicating universal neuron behavior, while middle layers exhibit diverse CVs with many highly specialized neurons. This layer-wise heterogeneity requires layer-specific expert configurations rather than uniform MoE conversion.


\section{Methodology}
\label{sec:methodology}

This section introduces \textbf{ExpertWeaver}, a training-free framework that converts dense LLMs into efficient MoE architectures through a three-stage process: a) Capturing multi-task GLU activation patterns for each layer; b) Leveraging the CVs of these activations to determine the layer-specific ratio of shared to routed experts; c) Constructing shared experts from universal neurons, clustering specialized neurons into routed experts, and building the MoE router.

\subsection{Capturing Multi-Task Gating Activation Patterns}
\label{sec:step1}
A standard SwiGLU-based FFN layer \citep{shazeer2020glu} consists of three weight matrices: $W_{\text{gate}} \in \mathbb{R}^{d_{\text{model}} \times d_{\text{ffn}}}$, $W_{\text{up}} \in \mathbb{R}^{d_{\text{model}} \times d_{\text{ffn}}}$, and $W_{\text{down}} \in \mathbb{R}^{d_{\text{ffn}} \times d_{\text{model}}}$.
Here, $d_{\text{model}}$ is the model's hidden dimension and $d_{\text{ffn}}$ is the dimension of the intermediate layer. Layer indices are omitted for notational simplicity. 
We define a \textbf{neuron slice} $\vs_j$ for each neuron $j \in \{1, \dots, d_{\text{ffn}}\}$ as:
$$
\vs_j = ( (\mW_{\text{gate}})_{:, j}, (\mW_{\text{up}})_{:, j}, (\mW_{\text{down}})_{j, :} ).
$$
The neuron slices are independent of each other. ExpertWeaver aims to partition the neuron slices $\{\mathbf{s}_j\}_{j=1}^{d_{\text{ffn}}}$ into different expert groups.

As established in \S\ref{sec:motivation}, GLU activation patterns are a rich source of information about the model's inherent structure. We use a multi-task calibration set $\mathcal{D}_{\text{calib}}$ from Flan-v2 (42 tasks, 5 samples each) to capture robust activation signals. For each sample $j$ in $\mathcal{D}_{\text{calib}}$, we compute the token-averaged gate activation for neuron $i$ as $a_{ij} = \text{mean}(\text{Swish}(\bm{x}_j \mW_{\text{gate}}))_i$, where $\bm{x}_j$ represents all tokens in sample $j$. This gives the activation profile of neuron $i$ as $\va_i = [a_{i1}, a_{i2}, \dots, a_{iM}]$ with $M = |\mathcal{D}_{\text{calib}}|$.
We collect all activation profiles into a matrix $\mA = [\va_1, \va_2, \dots, \va_{d_{\text{ffn}}}]$, which serves as the primary signal for our subsequent process.

\begin{figure}
    \centering
\includegraphics[width=0.9\linewidth]{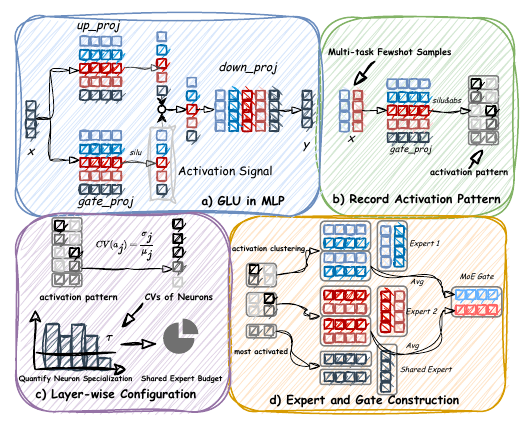}
    \caption{\textbf{The ExpertWeaver Framework.} a) The GLU in the MLP layer contains three weight matrices, where the same color denotes corresponding neuron slices. b) Neuron activation patterns are captured using a multi-task calibration dataset. c) The CVs are computed to determine the budget for shared vs. routed experts. (d) Neurons are clustered according to their activation patterns to form one shared expert and multiple routed experts.}
    \label{fig:method}
\end{figure}
\subsection{Layer-Aware Expert Allocation}
\label{sec:expert_allocation}
Building on our observation that the ratio of universal to specialized neurons varies across layers (\textbf{Takeaway 4}), we propose a layer-adaptive allocation strategy that determines the shared expert ratio for each layer.

\paragraph{Quantifying Layer-wise Neural Specialization.} First, we quantify the functional specialization of each layer $\ell$. We compute the CV based on each neuron's activation pattern $\bm{A}_\ell$. A high CV indicates that a neuron is highly specialized, activating only for specific inputs. We then define the layer's overall specialization ratio, $r_\ell$, as the fraction of neurons whose CV exceeds a CV threshold ($\tau$) used to identify specialized neurons:
\begin{equation}
    \label{eq:specialization_ratio}
    r_\ell = \frac{1}{d_{\text{ffn}}}\sum_{j=1}^{d_{\text{ffn}}}\mathbb{I}[CV_j > \tau],
\end{equation}
where $\mathbb{I}[\cdot]$ is the indicator function and $\tau$ is a specialization threshold.

\paragraph{Dynamic Allocation of Shared Expert Size.} Next, we use this specialization ratio to determine the proportion of neurons, $\alpha_\ell$, to be allocated to the shared expert in that layer. The core principle is that more specialized layers (higher $r_\ell$) require a smaller shared expert. We compute $\alpha_\ell$ using a linear mapping:
\begin{equation}
    \label{eq:shared_expert_ratio}
    \alpha_\ell = \alpha_{\max} - (\alpha_{\max} - \alpha_{\min}) \cdot r_\ell,
\end{equation}
where $\alpha_{\min}$ and $\alpha_{\max}$ define the bounds for the shared expert neuron ratio.
In our framework, the total $d_{\text{ffn}}$ neurons are distributed among $N_e$ experts, giving each expert a fixed capacity of $d_{\text{expert}} = d_{\text{ffn}} / N_e$ neurons. The number of shared experts, $N_{se, \ell}$, is then determined by the number of full experts that can be formed from the allocated shared neurons by:
\begin{equation}
    d_{s, \ell} = \text{round}(\alpha_\ell \cdot d_{\text{ffn}}), \quad \quad N_{se, \ell} = \text{round}(d_{s, \ell} / d_{\text{expert}}).
\end{equation}
The remaining $N_{re,l} = N_e - N_{se, \ell}$ experts are designated as routed experts. For each input token, we activate a total of $k$ experts according to the required sparsity. This includes activating all $N_{se, \ell}$ shared experts, plus the top $k - N_{se, \ell}$ routed experts as determined by the router.

\subsection{MoE Layer Construction}
\label{sec:step3}
Once the layer-specific allocations of shared experts ($N_{se, \ell}$) and routed experts ($N_{re,l}$) are determined, we proceed to partition the FFN's neurons into \textit{shared} and \textit{routed} experts.


\paragraph{Constructing Shared Experts.}
The shared expert is formed from the most universally active neurons. We select the top $d_{expert}\cdot N_{se, \ell}$ neurons with the highest absolute average activation scores to form the shared neuron pool, indexed by $\mathcal{I}_s$. These neuron slices are then concatenated to form the weight matrices of a single, consolidated shared expert. Specifically, the weight matrices for the shared expert are constructed by concatenating the corresponding neural slices from the original dense layer:
\begin{align}
    \mW_{\text{gate/up}}^{(\text{s})} &= \underset{j \in \mathcal{I}_s}{\text{CONCAT}}((\mW_{\text{gate/up}})_{:,j}) \\
    \mW_{\text{down}}^{(\text{s})} &= \underset{j \in \mathcal{I}_s}{\text{CONCAT}}((\mW_{\text{down}})_{j,:})
\end{align}
This shared expert is always activated, capturing and consolidating common knowledge across varying contexts.


\paragraph{Constructing Routed Experts.}
The remaining $N_{re, \ell} \cdot d_{\text{expert}}$ neurons, which form the specialized pool (indexed by $\mathcal{I}_{-s}$), are partitioned into $N_{re, \ell}$ routed experts. To group neurons that are frequently co-activated into the same expert, we employ a balanced K-Means clustering algorithm~\citep{malinen2014balanced} on their activation pattern vectors $\{\bm{a}_i\}_{i \in \mathcal{I}_{-s}}$ (see Appendix~\ref{app:balanced_kmeans} for details of balanced K-Means). This process partitions the set of specialized neurons into $N_{re, \ell}$ disjoint clusters $\{\mathcal{C}_1, \dots, \mathcal{C}_{N_{re, \ell}}\}$, where each cluster contains exactly $d_{\text{expert}}$ neurons and corresponds to a single routed expert.
The weight matrices for the $i$-th expert are formed by concatenating the weights of all neurons in cluster $\mathcal{C}_i$:
\begin{align}
    \label{eq:routed_expert_compact}
    \mW_{\text{gate/up}}^{(i)} &= \underset{j \in \mathcal{C}_i}{\text{CONCAT}}(\mW_{\text{gate/up}, j}) \\
    \mW_{\text{down}}^{(i)} &= \underset{j \in \mathcal{C}_i}{\text{CONCAT}}(\mW_{\text{down}, j})
    \end{align}
    
\paragraph{Constructing the MoE Router.} We construct our MoE router in a training-free manner by leveraging the clustering structure established in the previous step. The key insight is that the original gating vector of each neuron, $(\mW_{\text{gate}})_{:, j}$, controls that neuron's activation patterns. Therefore, the centroid of these vectors within each expert cluster naturally captures that expert's representative activation behavior.
Thus, we construct the router $\mW_{\text{router}} \in \mathbb{R}^{d_{\text{model}} \times N_{e, \ell}}$ by calculating a representative gating vector for each routed expert $i$ through averaging the gating vector of all neurons assigned to that cluster:
\begin{equation}
    \resizebox{.95\linewidth}{!}{$\displaystyle
\bar{\vw}_{\text{gate}}^{(i)} = \frac{1}{|\mathcal{C}_i|} \sum_{j \in \mathcal{C}_i} (\mW_{\text{gate}})_{:, j}, \quad \mW_{\text{router}} = \left[ \bar{\vw}_{\text{gate}}^{(1)}, \dots, \bar{\vw}_{\text{gate}}^{(N_{e,\ell})} \right].
$}
\end{equation}
By directly using the original gating weights, this approach constructs the router without training while preserving the neuron activation patterns captured during pretraining.


\subsection{ExpertWeaver in Practice: Dynamic Structural Pruning and Downcycling}
\label{sec:practice}

\subsubsection{Training-free Dynamic Structural Pruner}
When operating at low sparsity levels without requiring additional training, ExpertWeaver can be viewed as a dynamic structural pruning method.
For any given input $\bm{x}$, the router selects the top-$k$ routed experts based on the logits from the reconstructed router, $\mT(\bm{x}) = \text{TopK}(\bm{x} \mW_g)$. The final output is the direct sum of the outputs from the always-active shared experts and these selected routed experts:
\begin{equation}
    \vy = \mE_{\text{shared}}(\vx) + \sum_{i \in \mT(\vx)} \mE_i(\bm{x})
\end{equation}
It is worth noting that, instead of using a weighted combination of expert outputs, our gate acts as a structured pruning mechanism, dynamically selecting which neurons to exclude from the forward pass. This preserves the integrity of the original neuron-level computations from the pretrained model, as the GLU mechanism within each activated expert still governs the precise activation values.

\subsubsection{Downcycling Dense LLMs into MoEs}
Model downcycling aims to convert large, pretrained dense models into computationally efficient MoEs, using the dense model's weights as a superior initialization to avoid the prohibitive costs of training from scratch. ExpertWeaver performs \textbf{downcycling} by converting dense models into sparse MoE architectures, followed by continued Pretraining (CPT) for further optimization.

\paragraph{Initialization.}
Since the sparsity in downcycling is often higher and configured according to downstream requirements, we use a \textit{fixed, uniform shared expert ratio} across all layers. This provides a robust and well-structured starting point for the subsequent training phase.

\paragraph{Continued Pretraining.}
During CPT, we switch to a standard softmax router to provide greater optimization flexibility and enable experts to learn distinct specializations. The softmax router computes gating weights $\vg(\vx) = \text{Softmax}(\vx \mW_{\text{router}})$. The layer's output is a weighted combination of the shared expert and the top-k routed experts:
\begin{equation}
    \vy = \mE_{\text{shared}}(\vx) + \sum_{i \in \text{TopK}(\vg(\vx))} \vg(\vx)_i \cdot \mE_i(\vx).
\end{equation}
The total loss function consists of two components: the next-token prediction loss ($\mathcal{L}_{\text{NTP}}$) and an auxiliary load-balancing loss ($\mathcal{L}_{\text{LB}}$) to encourage a balanced distribution of tokens across experts.
\begin{equation}
    \mathcal{L}_{\text{total}} = \mathcal{L}_{\text{NTP}} + \lambda \mathcal{L}_{\text{LB}}; \quad \mathcal{L}_{\text{LB}} = N_{e,\ell} \cdot \sum_{i=1}^{N_{e,\ell}} f_i \cdot P_i,
\end{equation}
where $f_i$ is the fraction of tokens routed to expert $i$ and $P_i$ is its average router probability within a batch.

\section{Experiments}
We evaluate ExpertWeaver in the following two scenarios. First, as a training-free dynamic structured pruner for low-sparsity settings, we benchmark it against some training-free baselines in Section~\ref{sec:pruning_exp}. Second, as an initialization strategy for model downcycling in higher-sparsity settings, we demonstrate its advantages by comparing our resulting MoE models against similarly scaled baselines in Section~\ref{sec:downcycling_exp}.

\subsection{ExpertWeaver for Dynamic Structural Pruning}
\label{sec:pruning_exp}

\paragraph{Experimental Setup.}
\begin{table}
    \centering
    \caption{\textbf{Comparison with structured pruning methods under 25\% sparsity.}}
\label{tab:structured_pruning}
    \resizebox{\linewidth}{!}{%
\begin{tabular}{lcccccc}
    \toprule
    Method & MMLU(5) & HellaSwag(10) & ARC-e & ARC-c & PiQA & Avg. \\
    \midrule
    \multicolumn{7}{c}{\small LLaMA3-8B, 25\% sparsity} \\
    \hline
    Dense         & 65.3 & 82.1 & 77.9 & 57.9 & 80.8 & 72.8 \\
    LLM-Pruner    & 24.2 & 51.3 & 58.9 & 32.4 & 74.4 & 48.2 \\
    FLAP          & 33.4 & 48.0 & 50.0 & 29.3 & 68.3 & 45.8 \\
    CMoE      & 41.6 & 65.9 & 63.1 & 41.5 & 73.9 & 57.2 \\
    \rowcolor{gray!10}
    ExpertWeaver   & \textbf{47.0} & \textbf{69.8} & \textbf{64.4} & \textbf{44.3} & \textbf{76.3} & \textbf{60.4} \\
    \midrule
    \multicolumn{7}{c}{\small Qwen2.5-7B, 25\% sparsity} \\
    \hline
    Dense         & 74.2 & 80.3 & 77.8 & 63.8 & 80.0 & 75.2 \\
    LLM-Pruner    & 55.9 & 72.2 & 71.0 & 49.1 & \textbf{77.0} & 65.0 \\
    FLAP          & 54.7 & 58.5 & 67.3 & 42.2 & 70.8 & 58.7 \\
    \rowcolor{gray!10}
    ExpertWeaver  & \textbf{61.6} & \textbf{72.3} & \textbf{71.5} & \textbf{53.5} & 76.3 & \textbf{67.0} \\

    \bottomrule
\end{tabular}
    }
\end{table}

We benchmark ExpertWeaver against training-free baselines including FLAP \citep{an2024fluctuation}, LLM-Pruner \citep{ma2023llmpruner}, and CMoE \citep{pei2025cmoe} on Qwen2.5-7B and Llama3-8B models. Since CMoE is only implemented for Llama-series models, we only compared it on Llama3-8B. Following previous settings \citep{ma2023llmpruner}, we set the target sparsity level to 25\% for all methods. Model performance is assessed on five widely-used benchmarks: MMLU, HellaSwag, ARC-e, ARC-c, and PIQA. For ExpertWeaver, we use our default hyperparameters: a shared expert ratio with $\alpha_{\min}=0.2$ and $\alpha_{\max}=0.7$, a specialization threshold of $\tau=0.6$, and an expert granularity of 64. The specific layer-wise allocation between shared and routed experts is shown in Appendix~\ref{app:layer_wise_config} and the detailed ablation studies on hyperparameters is shown in Appendix~\ref{app:ablation_studies}. To construct an effective calibration set that captures diverse multi-task activation patterns, we sample 5 instances from each of the 42 tasks in the Flan-v2 dataset, which spans 10 distinct task clusters (details in Appendix~\ref{app:flan_details}). 

\paragraph{Main Results.}
Table~\ref{tab:structured_pruning} summarizes the performance comparison between ExpertWeaver and other training-free structural pruning baselines at 25\% sparsity. From these results, we observe the following key findings:
(1) ExpertWeaver consistently outperforms all competing approaches, achieving a 5.6\% relative improvement over CMoE on LLaMA3-8B and a 3.1\% advantage over LLM-Pruner on Qwen2.5-7B.
(2) Dense-to-MoE methods significantly surpass static pruning approaches like FLAP and LLM-Pruner. Rather than permanently removing parameters, these methods achieve sparsity by selectively activating parameter subsets based on input context, avoiding inevitable knowledge loss.
(3) ExpertWeaver outperforms CMoE for three key reasons. First, our comprehensive analysis of GLU activation patterns ensures that the original activation structure remains intact during conversion. Second, our multi-task calibration set enables better expert clustering by capturing diverse neural co-activation patterns (ablation in Appendix~\ref{app:ablation_calibration}). Third, our layer-specific configuration allows adaptive MoE construction tailored to each layer's specialization characteristics. We also compared our method against other structural pruning techniques on more complex tasks and evaluated its performance on a reasoning model, as detailed in Appendix~\ref{app:dynamic_pruning}.

\subsection{Downcycling with ExpertWeaver.}
\label{sec:downcycling_exp}
\paragraph{Experiment Setup.}
To evaluate ExpertWeaver as a downcycling strategy, we initialize two MoE variants. The first, ExpertWeaver (Qwen2.5-7B), is initialized from the dense Qwen2.5-7B model. It divides the MLP layers into 62 routed experts and 2 shared experts, activating 14 routed and 2 shared experts per token (7B total parameters, 3.5B active). This model undergoes CPT for 200B tokens on the FineWeb-Edu dataset~\citep{penedo2024fineweb}. For a fairer comparison with the OLMoE baseline, we also initialize a second variant, ExpertWeaver (OLMo-7B), from the OLMo-7B base model. This variant is continually pretrained for 200B tokens on the same dataset used for OLMoE. The performance of both models is evaluated on a wide range of tasks, including MMLU, HellaSwag, ARC-e, ARC-c, PIQA, WinoGrande, LogiQA, and SciQ, and compared against dense and MoE baselines with similar parameter counts and training budgets (see Appendix~\ref{app:baselines} for details).
Following CPT, we conduct a two-stage supervised fine-tuning (SFT) process, similar to Llama-MoE-V2~\citep{qu2024llama}, focusing first on general conversational abilities and then on code and math skills. The resulting instruction-tuned model is then compared with other instruction-tuned baselines on MMLU, ARC-c, GSM8K, HumanEval, and IFEval. Further training details are available in Appendix~\ref{app:training_configs} and the comparation results is shown in Appendix~\ref{app:sft_perf}.

\begin{figure*}[ht]
    \centering
    \includegraphics[width=.9\linewidth]{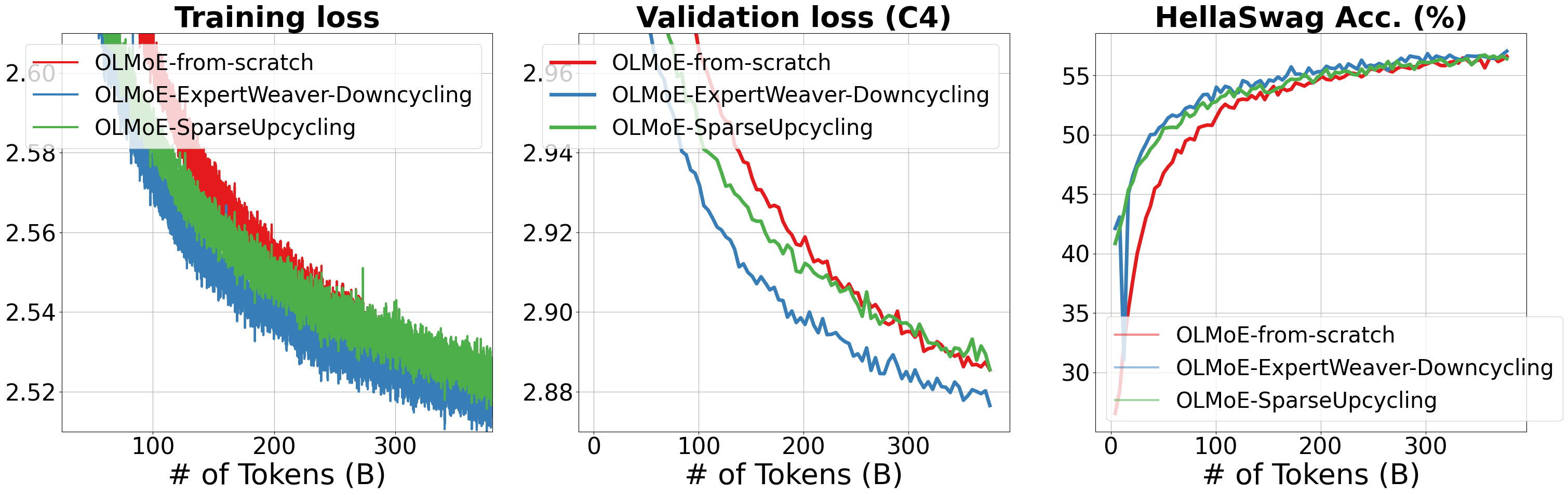}
    \caption{
    {\textbf{Comparison of Downcycling, Upcycling, and From-Scratch Training.} Comparison of training loss, evaluation loss, and downstream task performance using the same OLMoE model configuration under three different MoE initialization paradigms.}}
    \label{fig:olmo_downcycling_comparison}
\end{figure*}

\begin{table}[t]
\caption{\textbf{Downcycling Performance Comparison.} Best results are in bold, second-best are underlined. Models in gray (Qwen2.5-\{7B, 3B, 1.5B\}, Llama-3.2-3B, and OLMoE) are for context, as they were trained on significantly more data. Our ExpertWeaver models (OLMo-based and Qwen2.5-based) were trained on 200B tokens. The OLMo-based version has 1B active parameters (7B-A1B), and the Qwen2.5-based version has 3.5B active parameters (7B-A3.5B). We compare against the 500B token checkpoint (*) of OLMoE. Blank entries for OpenMoE indicate unavailable results.}
\label{tab:downcycling_results}
\centering
\resizebox{\linewidth}{!}{
\begin{tabular}{l|cccccccc|c}
\toprule
\textbf{Model} & \textbf{MMLU(5)} & \textbf{HellaSwag(10)} & \textbf{ARC-e} & \textbf{ARC-c(25)} & \textbf{PIQA} & \textbf{WinoGrande} & \textbf{LogiQA} & \textbf{SciQ} & \textbf{Average} \\
\midrule
\multicolumn{10}{l}{\textit{Dense Models}} \\
\textcolor{gray}{Qwen2.5-7B}         & \textcolor{gray}{74.1} & \textcolor{gray}{80.2} & \textcolor{gray}{77.5} & \textcolor{gray}{63.7} & \textcolor{gray}{79.7} & \textcolor{gray}{73.2} & \textcolor{gray}{36.4} & \textcolor{gray}{95.2} & \textcolor{gray}{72.5} \\

\textcolor{gray}{Qwen2.5-3B}         & \textcolor{gray}{65.6} & \textcolor{gray}{74.6} & \textcolor{gray}{73.9} & \textcolor{gray}{56.5} & \textcolor{gray}{78.8} & \textcolor{gray}{68.1} & \textcolor{gray}{33.5} & \textcolor{gray}{95.2} & \textcolor{gray}{68.3} \\
\textcolor{gray}{Qwen2.5-1.5B}       & \textcolor{gray}{60.9} & \textcolor{gray}{68.0} & \textcolor{gray}{72.5} & \textcolor{gray}{54.9} & \textcolor{gray}{75.9} & \textcolor{gray}{63.8} & \textcolor{gray}{31.9} & \textcolor{gray}{93.4} & \textcolor{gray}{65.2} \\
\textcolor{gray}{Llama-3.2-3B}       & \textcolor{gray}{56.1} & \textcolor{gray}{76.4} & \textcolor{gray}{71.6} & \textcolor{gray}{50.5} & \textcolor{gray}{77.4} & \textcolor{gray}{69.9} & \textcolor{gray}{30.6} & \textcolor{gray}{92.7} & \textcolor{gray}{65.7} \\
OLMo-7B       & 30.7 & 77.1 & 68.7 & 45.1 & 79.6 & 66.5 & 27.5 & 88.6 & 60.8 \\
OPT-2.7B           & 25.8 & 61.4 & 54.4 & 34.0 & 74.8 & 60.8 & 25.8 & 78.9 & 52.0 \\
Pythia-2.8B        & 26.8 & 60.7 & 58.8 & 36.7 & 73.6 & 59.6 & 28.1 & 83.2 & 53.4 \\
INCITE-Base-3B     & 27.2 & 64.7 & 61.7 & 40.3 & 73.9 & 63.5 & 27.5 & 85.6 & 55.6 \\
Open-LLaMA-3B-v2   & 26.8 & 71.4 & 63.3 & 40.1 & \underline{77.9} & 63.1 & 28.1 & 88.0 & 57.3 \\
Sheared-LLaMA-2.7B & 27.3 & 71.0 & 63.3 & 41.6 & 76.9 & 65.0 & 28.3 & 87.5 & 57.6 \\
Gemma-2-2b         & \textbf{53.0} & 69.0 & 36.9 & 52.6 & 67.5 & 51.9 & 22.7 & 75.8 & 53.7 \\
SmolLM2-1.7B       & \underline{50.4} & 72.6 & \textbf{73.4} & \underline{53.2} & 76.0 & \textbf{65.8} & \textbf{30.1} & 84.3 & \underline{63.2} \\
\midrule
\multicolumn{10}{l}{\textit{MoE Models}} \\
LLaMA-MoE-v1-3.5B   & 26.8 & \underline{73.3} & 65.6 & 44.2 & \underline{77.9} & \underline{65.5} & \underline{29.7} & 87.6 & 58.8 \\
OpenMoE-3B-9B       & - & 56.5 & 50.6   & 33.3 & 65.7 & 51.9   & -    & - & -    \\
\textcolor{gray}{OLMoE-1B-7B} & \textcolor{gray}{53.8} & \textcolor{gray}{79.6} & \textcolor{gray}{76.3} & \textcolor{gray}{55.6} & \textcolor{gray}{80.1} & \textcolor{gray}{68.4} & \textcolor{gray}{29.3} & \textcolor{gray}{94.9} & \textcolor{gray}{67.2} \\
{OLMoE-1B-7B*} & 28.4 & 70.2 & 71.0 & 43.9 & 77.1 & 63.5 & 28.7 & \underline{88.5} & 58.9 \\
LLaMA-MoE-v2-3.5B   & 40.9 & 53.7 & 57.0 & 40.2 & 67.9 & 56.1 & 30.7 & \textbf{88.8} & 54.4 \\
\rowcolor{gray!10}
ExpertWeaver-E64-A14-S2\scriptsize{(OLMo-7B)} & 45.0 & 61.2 & 69.3 & 38.8 & 74.5 & 62.1 & 28.5 & 91.8 & 58.9 \\
\rowcolor{gray!10}
ExpertWeaver-E64-A14-S2\scriptsize{(Qwen2.5-7B)} & 45.6 & \textbf{73.7} & \underline{72.4} & \textbf{56.3} & \textbf{78.0} & 65.3 & 29.0 & 87.7 & \textbf{63.5} \\
\bottomrule
\end{tabular}
}
\end{table}

\paragraph{Main Results.}
Table~\ref{tab:downcycling_results} presents the results of comparing ExpertWeaver with other models of comparable parameters and training budgets, yielding the following observations. (1) Among models with comparable training budgets and parameter counts, our ExpertWeaver{\scriptsize(Qwen2.5-7B)} achieves the best average performance (63.5), outperforming the strongest MoE baseline (OLMoE-1B-7B*) by 4.6 points. (2) To ensure a fair comparison, we also downcycled OLMo-7B. After just 200B tokens of continued pretraining, our ExpertWeaver{\scriptsize(OLMo-7B)} achieves a score of 58.9. This performance represents 97.35\% of the original OLMo-7B model's performance (60.5). Notably, this result not only nears the performance of the original model but also matches the performance of the OLMoE-1B-7B* model, which was trained from scratch on more than double the data (500B tokens). While a slight difference in scores may be attributed, in part, to the potentially lower quality of the Dolma dataset used by OLMo compared to OLMoE-Mix~\citep{muennighoff2024olmoe}, this result strongly highlights the effectiveness of our ExpertWeaver method. It demonstrates the capacity to efficiently downcycle a large dense model into a high-performing MoE architecture with only a minimal amount of continued training data. (3) Compared to its dense counterpart, ExpertWeaver{\scriptsize(Qwen2.5-7B)} retains 87.6\% of the base Qwen2.5-7B's performance while activating only a quarter of the MLP parameters. This remarkable performance retention, achieved with a modest 200B tokens of training, strongly validates the effectiveness of our downcycling strategy. Furthermore, when compared to dense models with similarly activated parameter counts like Qwen2.5-3B and Llama-3.2-3B, our model achieves 93.0\% and 96.7\% of their respective performance. Given that these models were trained on massive datasets (18T and 9T tokens, respectively), this result further demonstrates that our downcycling method is a highly efficient path to creating performant MoE models.

\begin{figure}[ht]
    \centering
      \includegraphics[width=.8\linewidth]{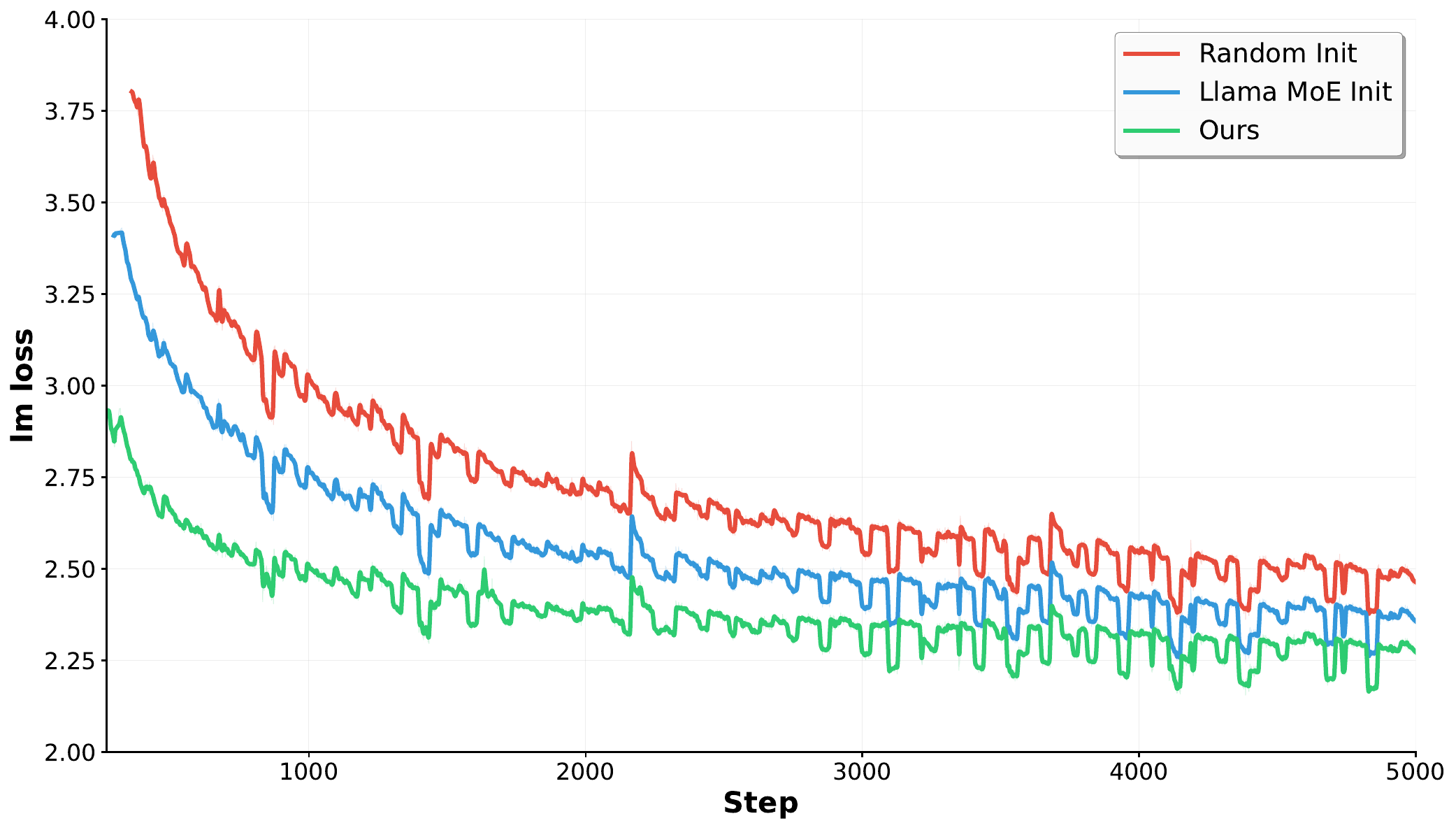}
      \caption{\textbf{Training Loss Comparison with Different MoE Initialization Strategies.}}
    \label{fig:training_loss_comparison}
    \end{figure}    
\paragraph{Compare with other MoE initialization methods.}
Figure \ref{fig:training_loss_comparison} shows the training loss for the first 5,000 steps (approximately 20B tokens), highlighting the effectiveness of our ExpertWeaver as a downcycling strategy. Compared to both random and Llama-MoE initializations, our method consistently yields a lower training loss throughout the entire training process. This demonstrates that ExpertWeaver provides a more effective starting point for the MoE, enabling faster convergence and superior final performance. The persistent gap between our loss curve (green) and the others validates that our strategy better preserves the foundational knowledge of the dense model during conversion.

\paragraph{A Direct Comparison of Downcycling, Upcycling, and From-Scratch Training.}  
To provide a fairer comparison and demonstrate the effectiveness of ExpertWeaver, we used OLMo-1.3B (pre-trained on 1T tokens) as the base model for our downcycling process. For direct contrast, we also performed sparse upcycling~\citep{muennighoff2024olmoe}, training a 575M OLMo model for 1T tokens to yield a 1.3B OLMoE model (with 676M activated parameters). Both our downcycling method and the upcycling approach are compared against a baseline trained from scratch. Detailed model configurations are provided in Appendix~\ref{app:model_arch}.
As illustrated in Figure~\ref{fig:olmo_downcycling_comparison}, which plots the training loss, evaluation loss, and downstream task performance, the results lead to several key observations. Initially, both the downcycling and upcycling strategies exhibit faster convergence compared to the from-scratch baseline. However, over a longer training period, the performance of the upcycled model regresses toward that of the from-scratch baseline. In contrast, our downcycling approach with ExpertWeaver consistently achieves the best performance and convergence, maintaining its superiority throughout the entire training process. By inheriting a richer feature space from a larger model and avoiding parameter duplication, downcycling offers more diverse experts and a higher optimization ceiling than weight-duplicating upcycling, which risks local optima.


\section{Conclusion}
In this paper, we investigate the problem of converting dense models into high-performance MoEs. We show that GLU activation patterns provide a rich source for identifying latent neuron specialization, which naturally enables effective expert construction. Based on this observation, we introduce ExpertWeaver, a training-free method that partitions neurons into shared and routed experts in a layer-wise manner based on their activation patterns. Experiments demonstrate that ExpertWeaver significantly outperforms existing methods in both zero-shot pruning for inference efficiency and MoE initialization for model downcycling. 


\bibliography{example_paper}
\bibliographystyle{icml2026}

\onecolumn
\appendix
\section{Impact Statement}
ExpertWeaver is a training-free method to convert pretrained dense LLMs into sparse MoE models using GLU activation patterns.
By activating fewer parameters per token, it can reduce inference compute and energy, potentially lowering deployment cost and environmental impact.
As with other efficiency improvements, cheaper inference may also enable broader misuse (e.g., spam or misinformation), and the conversion does not itself address issues such as bias or unsafe generation.
We recommend standard safety evaluation and monitoring when deploying converted models.

\section{Ablation Studies}
\label{app:ablation_studies}
\begin{figure}[h]
    \centering
      \includegraphics[width=.65\linewidth]{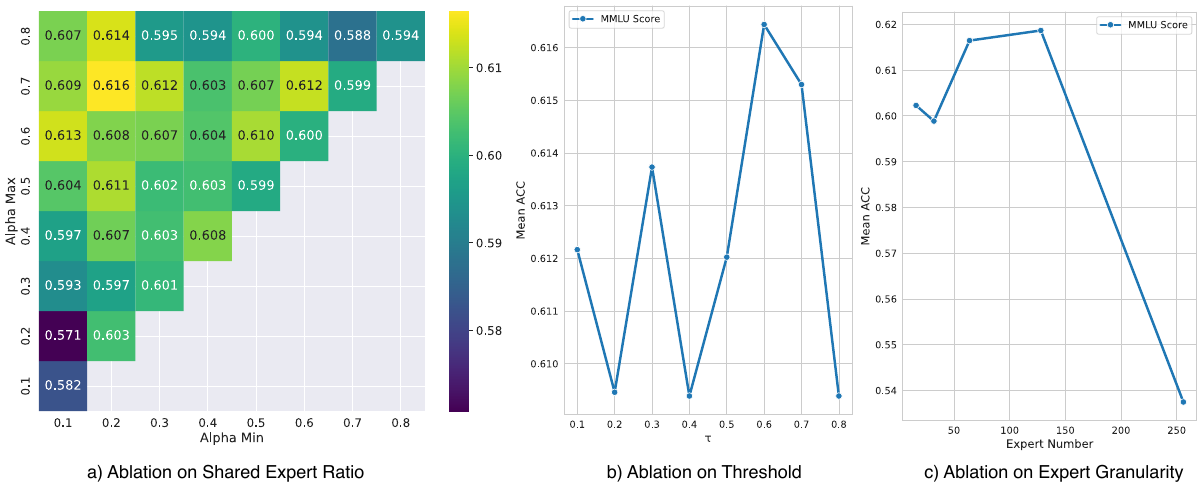}
        \caption{\textbf{Ablation studies on key hyperparameters of ExpertWeaver.} a) MMLU performance heatmap for different shared expert ratios, controlled by the hyperparameters $\alpha_{\min}$ and $\alpha_{\max}$ in Eq.~\ref{eq:shared_expert_ratio}. The diagonal where $\alpha_{\min} = \alpha_{\max}$ represents a static configuration with a uniform ratio across all layers. b) MMLU performance varying specialization threshold $\tau$ from Eq.~\ref{eq:specialization_ratio}. c) MMLU performance varying expert granularity.}
    \label{fig:ablation}
    \end{figure}
As shown in Figure~\ref{fig:ablation}, we conduct a series of ablation studies to investigate the impact of key hyperparameters. 
    
\textbf{a) Shared Expert Ratio:} The heatmap in Figure~\ref{fig:ablation}a) explores the impact of the shared expert ratio, controlled by $\alpha_{\min}$ and $\alpha_{\max}$ from Eq.~\ref{eq:shared_expert_ratio}. The results reveal that our layer-aware dynamic allocation strategy significantly outperforms static configurations. The diagonal, where $\alpha_{\min} = \alpha_{\max}$, represents a static setting with a uniform ratio across all layers, and its performance is lower than that of the off-diagonal regions. The optimal performance is achieved within the range of $\alpha_{\min} \in [0.2, 0.4]$ and $\alpha_{\max} \in [0.5, 0.7]$, with the peak at $(0.2, 0.7)$. We adopted $\alpha_{\min}=0.2$ and $\alpha_{\max}=0.7$ as our default setting.  \textbf{b) Specialization Threshold:} We study the impact of the specialization threshold $\tau$ from Eq.~\ref{eq:specialization_ratio}, which determines whether a neuron is classified as specialized based on its CV score. The model demonstrates robust performance across a range of $\tau$ values, with optimal results achieved at $\tau=0.6$. We adopt $\tau=0.6$ as the default configuration for all experiments. \textbf{c) Expert Granularity:} The total number of experts significantly influences performance. The MMLU score peaks when the number of experts is around 64 and 128, declining with either too few or too many experts. This suggests an optimal granularity that balances functional diversity and specialization efficiency. To prioritize inference efficiency while maintaining strong performance, we adopt 64 as our default expert granularity.

\begin{table}[ht]
    \centering
    \caption{\textbf{Instruct Model Performance Comparison}}
    \label{tab:sft_performance}
    \resizebox{.75\linewidth}{!}{%
\begin{tabular}{lcccccc}
    \toprule
    Method            & MMLU(5)  & ARC-C(25) & GSM8K & HumanEval & IFEval & Avg.  \\
    \midrule
     LLaMA-MoE-3B-7B       & 28.24 & 44.03 & 4.62  & 12.02 & 28.10 & 23.40 \\
    OLMoE-1B-7B           & 53.79 & 55.63 & 40.94 & 40.48 & 35.49 & 45.27 \\
    LLaMA-MoE-v2      & 40.90 & 40.20 & 55.00 & \textbf{51.20}    & \textbf{36.00}  & 44.66 \\
    \rowcolor{gray!10}
    ExpertWeaver-Instruct       & \textbf{50.60} & \textbf{69.80} & \textbf{57.10} & 50.20     & 33.10  & \textbf{52.16} \\
    \bottomrule
\end{tabular}
    }
\end{table}

\section{SFT Performance}
\label{app:sft_perf}
As shown in Table~\ref{tab:sft_performance}, our instruction-tuned model, ExpertWeaver-Instruct, achieves superior performance against other MoE baselines. It achieves a top-ranking average score of 52.16, establishing a clear lead over comparable MoE models like OLMoE-1B-7B (45.27) and LLaMA-MoE-v2 (44.66). This result demonstrates the effectiveness of the ExpertWeaver methodology in creating a robust foundation for supervised fine-tuning. We also show the serving efficiency of the ExpertWeaver-Instruct model in Appendix~\ref{app:serving_efficiency}.


\section{Related Work}
\label{app:related_work}

\paragraph{Structural Pruning.}
Structural pruning improves LLM efficiency by removing entire components like neurons or attention heads. To avoid costly retraining, recent training-free methods use importance scores to identify and remove structures. For example, LLM-Pruner~\citep{ma2023llmpruner} analyzes gradients, while FLAP~\citep{an2024fluctuation} measures output feature stability. These methods perform \textit{static} pruning by permanently removing weights. In contrast, methods that convert dense models into MoEs, such as ToMoE~\citep{gao2025tomoe}, can be considered a form of \textit{dynamic} structural pruning. Since the goal of converting a dense model to an MoE aligns with that of structural pruning (reducing computational cost while preserving performance), we also categorize ExpertWeaver as a dynamic structural pruning method. It achieves this by dynamically selecting a sparse subset of neuron blocks for each token at inference time.

\paragraph{Model Upcycling.}
Model \textit{upcycling}~\citep{team2024qwen2,muennighoff2024olmoe,he2024upcycling,nakamura2025drop,komatsuzaki2022sparse} offers a cost-effective strategy for creating larger sparse models by merging smaller, pretrained dense models, thus avoiding the expense of from-scratch training. A common technique is to replicate the FFN layers from one or more dense models to form the experts of a new, larger MoE, followed by a fine-tuning phase to learn the routing logic. While effective, this paradigm faces two emerging challenges. First, the concentration of research and computational resources on frontier models means that the capabilities of state-of-the-art large models are advancing at a pace that smaller models struggle to match. Upcycling from smaller, less capable models may therefore not match the performance of downcycling from a larger, more advanced one. Second, because upcycling often relies on parameter duplication to initialize experts, it can lead to a lack of diversity that predisposes the model to fall into local optima during long-term optimization. In contrast, downcycling, as implemented by ExpertWeaver, leverages the rich, non-redundant internal structure of a single large model, providing a more robust foundation for sustained performance gains.

\paragraph{Model Downcycling.} Model \textit{downcycling} converts large pretrained dense models into computationally efficient MoEs, aiming to retain performance while gaining inference speed through FFN neuron partitioning. Moefication~\citep{zhang2022moefication} pioneered this field by splitting FFN parameters into functional partitions as experts with learned routers, but was originally designed for ReLU-based networks and struggles with modern GLU-based architectures. LTE~\citep{zheng2024learn} trains efficiency-aware models to amplify inherent activation sparsity through efficiency loss penalties, but requires substantial computational overhead during the training phase and does not directly leverage inherent model structures for expert creation. ToMoE~\citep{gao2025tomoe} uses dynamic structural pruning with frozen weights to discover expert structures via learned routing, but still requires substantial training overhead for the routing modules. Llama-MoE~\citep{zhu2024llama,qu2024llama} partitions FFN parameters through clustering followed by extensive continual pretraining with 200B tokens, but ignores the model's internal structural patterns, resulting in poor MoE performance that fails to preserve the original dense model's capabilities. CMoE~\citep{pei2025cmoe} achieves training-free conversion using balanced clustering with analytically constructed routers, yet lacks detailed activation signal analysis and employs fixed configurations across layers, leading to suboptimal expert partitioning that fails to capture layer-specific specializations. ExpertWeaver introduces a novel training-free dense-to-MoE technique that leverages intrinsic GLU activation patterns to form experts and construct routers, enabling both training-free applications and supporting efficient downcyling for further CPT.

\section{Comparision on Complex Tasks and Reasoning Models}
\label{app:dynamic_pruning}

To compare the capabilities of different methods in more complex scenarios, we first investigate the performance of Qwen2.5-7B on GSM8k and HumanEval under various sparsities. Furthermore, to assess performance on models specialized for reasoning, we explore the effectiveness of these methods on the DeepSeek-R1-Distill-Qwen-7B model, evaluating it on the GPQADiamond and LiveCodeBench benchmarks.

The results, presented in Table~\ref{tab:pruning_code_benchmarks}, reveal the limitations of static structural pruning on complex tasks. At 25\% sparsity on GSM8k, static methods like LLM-Pruner and FLAP almost completely fail; the dash (-) for LLM-Pruner indicates its accuracy dropped to zero. In stark contrast, ExpertWeaver, which retains all model parameters, maintains strong performance. This advantage holds at 12.5\% sparsity, where ExpertWeaver continues to significantly outperform static methods on both GSM8k and HumanEval. The same trend is observed on the specialized reasoning model, DeepSeek-R1-Distill-Qwen-7B, where ExpertWeaver again achieves the highest scores on GPQADiamond and LiveCodeBench. This demonstrates that ExpertWeaver's dynamic pruning approach is substantially more effective at preserving critical reasoning and coding capabilities, especially at higher sparsity levels where static methods suffer from irreversible information loss.

\begin{table*}[th]
\centering
\caption{\textbf{Training-Free Pruning on Code Benchmarks.} We compare ExpertWeaver against static pruning methods on several code-related benchmarks. The base models used for each benchmark are specified in the table.}
\label{tab:pruning_code_benchmarks}
\begin{tabular}{lrrrrcc}
\toprule
& \multicolumn{2}{c}{GSM8k} & \multicolumn{2}{c}{HumanEval} & GPQADiamond & LiveCodeBench \\
\cmidrule(lr){2-7}
\textbf{Base Model} & \multicolumn{4}{c}{Qwen2.5-7B} & \multicolumn{2}{c}{DeepSeek-R1-Distill-Qwen-7B} \\
\cmidrule(lr){2-5} \cmidrule(lr){6-7}
\textbf{Sparsity} & 25\% & 12.5\% & 25\% & 12.5\% & \multicolumn{2}{c}{12.5\%} \\
\midrule
LLM-Pruner & 2.0 & - & 14.6 & 10.4 & 33.3 & 6.1 \\
FLAP & 14.8 & - & 57.6 & 9.8 & 36.3 & 2.3 \\
ExpertWeaver & 34.9 & 18.9 & 64.6 & 32.9 & 37.3 & 16.8 \\
\bottomrule
\end{tabular}
\end{table*}

\section{Serving Efficiency}
\label{app:serving_efficiency}

To investigate the serving efficiency of our model, we conduct a comprehensive benchmark comparing ExpertWeaver (E64-A14-S2, derived from Qwen2.5-7B) with the dense Qwen2.5-7B model using the vLLM framework. All tests are run on a single GPU with `tensor\_parallel\_size=1' and GPU memory utilization set to 90\% to maximize the memory allocated to the KV Cache. We simulate a high-load scenario by sending 1024 random requests with an average input length of 512 tokens at an infinite rate, evaluating the models' peak performance at a maximum concurrency of 128 sequences. The evaluation spans three generations of NVIDIA GPUs—A100 (Ampere) and H100 (Hopper)—to ensure a comprehensive and fair assessment.

The results, presented in Table~\ref{tab:serving_metrics_gpus}, show ExpertWeaver demonstrates superior performance on both platforms. It achieves higher throughput (RPS, OTPS, and TTPS) and lower latency for both the first token (TTFT) and subsequent tokens (TPOT). This clear-cut advantage across all metrics confirms the inference efficiency of the ExpertWeaver.

\begin{table}[h]
\centering
\caption{\textbf{Inference throughput comparison across different GPU architectures.} RPS: Requests Per Second. OTPS: Output Tokens Per Second. TTPS: Total Tokens Per Second. TTFT: Time To First Token. TPOT: Time Per Output Token. ITL: Inter-Token Latency.}
\label{tab:serving_metrics_gpus}
\resizebox{.75\linewidth}{!}{
\begin{tabular}{llrrrrrr}
\toprule
\textbf{GPU} & \textbf{Method} & \textbf{RPS ↑} & \textbf{OTPS ↑} & \textbf{TTPS ↑} & \textbf{TTFT ↓} & \textbf{TPOT ↓} & \textbf{ITL ↓} \\
& & & & & \textbf{(ms)} & \textbf{(ms)} & \textbf{(ms)} \\
\midrule
\multirow{2}{*}{\textbf{A100}} & ExpertWeaver & 22.88 & 2928.86 & 14644.31 & 440.6 & 40.3 & 40.3 \\
& Qwen2.5-7B & 18.85 & 2413.22 & 12066.10 & 675.4 & 47.8 & 48.2 \\
\midrule
\multirow{2}{*}{\textbf{H100}} & ExpertWeaver & 44.04 & 5637.42 & 28187.10 & 542.2 & 9.8 & 9.8 \\
& Qwen2.5-7B & 41.41 & 5299.96 & 26499.78 & 1404.5 & 18.5 & 18.5 \\
\bottomrule
\end{tabular}
}
\end{table}

\section{Ablation Studies on the Calibration Set}
\label{app:ablation_calibration}
To investigate the impact of the calibration set on ExpertWeaver's performance, we conducted two ablation studies, with results presented in Table~\ref{tab:ablation_calibration}.

\paragraph{Impact of Data Quantity and Diversity.}
We first analyzed the sensitivity to the calibration set's size and diversity. We created several calibration sets by randomly sampling different proportions (25\%, 50\%, 75\%, and 100\%) of the tasks from our default Flan-v2 collection. For each selected task, we used 10 samples, meaning that as the proportion increases, both the number of tasks (diversity) and the total number of samples (quantity) grow. The results show that performance generally improves with a larger and more diverse calibration set, with the best result (67.0) achieved using 100\% of the tasks. Notably, using just 50\% of the tasks already yields a strong average score of 66.1, which is over 98\% of the final performance. This demonstrates that while diversity is beneficial, ExpertWeaver is data-efficient and does not require an excessively large calibration set to achieve robust performance.

\paragraph{Impact of Data Source.}
We also compared our default multi-task calibration set (Flan-v2) against a variant calibrated solely on a general-domain corpus (C4), denoted as ExpertWeaver$_{C4}$. The results clearly demonstrate the benefit of using a multi-task dataset. Our default ExpertWeaver achieves a superior average score of 67.0, outperforming ExpertWeaver$_{C4}$ (65.8). This suggests that capturing a wide range of activation patterns from diverse tasks is crucial for identifying a truly robust and generalizable functional structure within the dense model, leading to a more effective expert partition.

\begin{table*}[ht]
\caption{\textbf{Ablation Studies on the Calibration Set.} We analyze the impact of calibration set size, diversity, and source.}
\label{tab:ablation_calibration}
\centering
\begin{tabular}{lcccccc}
    \toprule
    Method & MMLU & HellaSwag(10) & ARC-e & ARC-c(25) & PiQA & Avg. \\
    \midrule
    \multicolumn{7}{c}{\small Qwen2.5-7B, 25\% sparsity} \\
    \midrule
    Dense         & 74.2 & 80.3 & 77.8 & 63.8 & 80.0 & 75.2 \\
    \midrule
    \multicolumn{7}{l}{\textit{Ablation on Data Source}} \\
    ExpertWeaver$_{Flan}$ (default)  & 61.6 & 72.3 & 71.5 & 53.5 & 76.3 & 67.0 \\ 
    ExpertWeaver$_{C4}$ & 61.2 & 71.6 & 72.1 & 49.1 & 74.8 & 65.8 \\
    \midrule
    \multicolumn{7}{l}{\textit{Ablation on Data Quantity \& Diversity}} \\
    ExpertWeaver$_{Flan 25\%}$  & 59.2 & 69.0 & 68.7 &  47.3 & 75.4 & 63.9 \\ 
    ExpertWeaver$_{Flan 50\%}$  & 60.6 & 71.2 & 71.5 &  51.7 & 75.5 & 66.1 \\ 
    ExpertWeaver$_{Flan 75\%}$  & 60.1 & 71.3 & 72.5 &  46.8 & 77.3 & 65.6 \\ 
    ExpertWeaver$_{Flan 100\%}$ & 61.6 & 72.3 & 71.5 & 53.5 & 76.3 & 67.0 \\ 
    \bottomrule
\end{tabular}
\end{table*}

\section{Details of the Calibration Set}
\label{app:flan_details}

To construct a diverse, multi-task calibration set for analyzing neuron activation patterns, we sampled from the Flan-v2 collection~\citep{chung2024scaling}. Flan-v2 is a large-scale dataset consisting of a mixture of publicly available NLP datasets that have been formatted into an instruction-tuning style. This diversity makes it an ideal source for a calibration set intended to capture a wide range of functional specializations.

For our calibration set, $\mathcal{D}_{\text{calib}}$, we selected a representative subset of 48 distinct tasks from 10 different task clusters within Flan-v2. For each of these 48 tasks, we randomly sampled 5 few-shot examples, resulting in a total of 240 samples in our calibration set. This carefully curated subset ensures that our analysis of neuron activation patterns is based on a broad and balanced distribution of tasks, from reading comprehension and summarization to commonsense reasoning and natural language inference.

Table \ref{tab:flan_tasks} lists the 10 task clusters and the 48 specific tasks used in our calibration set.

\begin{table*}[h]
\centering
\caption{Tasks from Flan-v2 used in the calibration set.}
\label{tab:flan_tasks}
\resizebox{\linewidth}{!}{%
\begin{tabular}{lll}
\toprule
\textbf{Task Cluster} & \textbf{Description} & \textbf{Datasets} \\
\midrule
Reading Comprehension & Answers questions based on provided passages. & squad\_v1, squad\_v2, drop, duorc, quac, record \\
Summarization & Creates a shorter version of a document. & xsum, cnn\_dailymail, samsum, multi\_news \\
Translation & Translates text across multiple languages. & wmt14\_en-fr, wmt14\_en-de, wmt14\_en-ro \\
Commonsense Reasoning & Understands everyday scenarios. & boolq, piqa, siqa, cosmos\_qa, hellaswag, winogrande \\
Natural Language Inference & Determines logical relationship between sentences. & mnli, qnli, rte, wnli, anli \\
Coreference Resolution & Identifies expressions referring to the same entity. & wsc, dpr, winogender \\
Sentiment Analysis & Determines sentiment polarity. & imdb, sentiment140, yelp\_polarity \\
Question Answering & Answers questions without external knowledge. & arc, openbookqa, race, trivia\_qa \\
Paraphrase Detection & Generates alternative phrasings of sentences. & mrpc, qqp \\
Structure-to-Text & Generates text from structured data. & common\_gen, e2e\_nlg, dart \\
\bottomrule
\end{tabular}%
}
\end{table*}

\section{Balanced K-Means Clustering}
\label{app:balanced_kmeans}

We employ balanced K-Means clustering to partition specialized neurons $\mathcal{I}_r$ into $N_{re, \ell}$ routed experts of equal capacity $d_{\text{expert}}$. Given activation pattern vectors $A_r = \{\bm{a}_i\}_{i \in \mathcal{I}_r}$, the algorithm finds clusters $\mathcal{C}_1, \dots, \mathcal{C}_{N_{re, \ell}}$ that solve:

\begin{equation}
\begin{aligned}
\min_{\mathcal{C}_1, \dots, \mathcal{C}_{N_{re, \ell}}} \quad & \sum_{k=1}^{N_{re, \ell}} \sum_{i \in \mathcal{C}_k} ||\bm{a}_i - \bm{\mu}_k||^2 \\
\text{s.t.} \quad & |\mathcal{C}_k| = d_{\text{expert}} \quad \forall k \in \{1, \dots, N_{re, \ell}\} \\
& \mathcal{C}_j \cap \mathcal{C}_k = \emptyset \quad \forall j \neq k \\
& \bigcup_{k=1}^{N_{re, \ell}} \mathcal{C}_k = \mathcal{I}_r
\end{aligned}
\end{equation}

where $\bm{\mu}_k$ is the centroid of cluster $\mathcal{C}_k$.

\textbf{Algorithm}:
\begin{enumerate}
    \item \textbf{Initialize}: Sample $N_{re, \ell}$ centroids $\bm{\mu}_1, \dots, \bm{\mu}_{N_{re, \ell}}$ from $A_r$.
    \item \textbf{Assign}: Solve the minimum-cost perfect matching problem with fixed centroids. We use a greedy approximation: iteratively assign neurons to the nearest available cluster slot.
    \item \textbf{Update}: Recompute centroids as:
    \begin{equation}
    \bm{\mu}_k \leftarrow \frac{1}{d_{\text{expert}}} \sum_{i \in \mathcal{C}_k} \bm{a}_i \quad \forall k
    \end{equation}
    \item \textbf{Iterate}: Repeat steps 2-3 until convergence.
\end{enumerate}

This ensures functionally coherent and structurally uniform experts for efficient MoE implementation.

\section{Details of the compared baselines}
\label{app:baselines}

\subsection{Baselines for Structured Pruning}
For the structured pruning evaluation, we compare ExpertWeaver with the following training-free methods:
\begin{itemize}
    \item \textbf{LLM-Pruner}~\citep{ma2023llmpruner} is a task-agnostic, training-free structural pruning method that identifies and removes redundant structures by analyzing gradient information and parameter magnitudes. It aims to preserve the model's generalization capabilities by focusing on the interconnectedness of model components.
    \item \textbf{FLAP}~\citep{an2024fluctuation} is a training-free pruning framework that prunes large language models at the FFN-layer level. It uses a metric based on output feature stability to identify and remove less important neurons, offering a computationally efficient alternative to methods that require gradient computation.
    \item \textbf{CMoE}~\citep{pei2025cmoe} is a training-free method that converts dense models into MoEs using balanced clustering. It constructs routers analytically but uses a fixed configuration across all layers, which may not capture layer-specific functional specializations.
\end{itemize}

\subsection{Baselines for Model Downcycling}
In the model downcycling experiments, we compare our ExpertWeaver-initialized model against a variety of both dense and MoE models with comparable parameter counts and training budgets.

\paragraph{Dense Models.} We include several strong, publicly available dense models as baselines:
\begin{itemize}
    \item \textbf{OPT-2.7B}~\citep{zhang2022opt}, \textbf{Pythia-2.8B}~\citep{biderman2023pythia}, \textbf{INCITE-Base-3B}~\citep{weber2024redpajama}, \textbf{Open-LLaMA-3B-v2}~\citep{geng2023openllama}, and \textbf{Sheared-LLaMA-2.7B}~\citep{xia2024sheared} are all well-established language models in the 2.7B-3B parameter range.
    \item \textbf{Gemma-2-2B}~\citep{team2024gemma} is a recent, highly-performant model from Google.
    \item \textbf{SmolLM2-1.7B}~\citep{allal2025smollm2} is another strong baseline known for its efficiency and performance at a smaller scale.
\end{itemize}

\paragraph{MoE Models.} We also compare against several existing MoE models:
\begin{itemize}
    \item \textbf{LLaMA-MoE-v1}~\citep{zhu2024llama} and \textbf{LLaMA-MoE-v2}~\citep{qu2024llama} are MoE variants of the LLaMA architecture. They are created by partitioning the FFN parameters of a dense model into experts and then applying extensive continued pretraining.
    \item \textbf{OpenMoE}~\citep{xue2024openmoe} is an open-source MoE model series.
    \item \textbf{OLMoE}~\citep{muennighoff2024olmoe} is a family of open language models with a MoE architecture, developed by AI2. We compare against a checkpoint trained on 500B tokens to ensure a fair comparison with our model's training budget.
\end{itemize}

\subsection{Baselines for Instruct Model Performance}
For evaluating the performance of our instruction-tuned model, \textbf{ExpertWeaver-Instruct}, we compare it against the following instruction-tuned MoE baselines:
\begin{itemize}
    \item \textbf{LLaMA-MoE-3B-7B} and \textbf{LLaMA-MoE-v2} are the instruction-tuned versions of the LLaMA-MoE models described above.
    \item \textbf{OLMoE-1B-7B} is the instruction-tuned version of the OLMoE model.
\end{itemize}

\section{Model Architecture for OLMo Downcycling Experiment}
\label{app:model_arch}
The model architecture for the OLMo Downcycling experiment is shown in Table~\ref{tab:model_config}.

\begin{table*}[h]
\centering
\caption{\textbf{Model Configurations of OLMo Downcycling Experiment}} 
\label{tab:model_config} 
\begin{tabular}{lccc}
\toprule
& \textbf{OLMo 575M} & \textbf{OLMo 1.3B} & \textbf{OLMoE 1.3B-A676M} \\
\midrule
Model Dimension & 2048 & 2048 & 2048 \\
FFN Dimension & 1024 & 8192 & 1024 \\
Attention Heads & 16 & 16 & 16 \\
Key/Value Heads & 16 & 16 & 16 \\
Layers & 16 & 16 & 16 \\
Vocabulary Size & 50280 & 50280 & 50280 \\
Weight Typing & True & True & True \\
Context Length & 4096 & 4096 & 4096 \\
Expert Granularity & - & - & 2 in 8 \\
\bottomrule
\end{tabular}
\end{table*}

\section{Training Configurations}
\label{app:training_configs}

This section provides the detailed configurations for both the continued pre-training (CPT) and supervised fine-tuning (SFT) phases.

\subsection{Continued Pre-training (CPT)}
The CPT phase was conducted on the ExpertWeaver-E64-A14-S2 model for 200 billion tokens using the FineWeb-Edu dataset \citep{penedo2024fineweb} using 128 H100 GPUs. Key hyperparameters are listed in Table~\ref{tab:cpt_hyperparams}. We use megatron-swift\footnote{\url{https://swift.readthedocs.io/en/latest/Megatron-SWIFT/Quick-start.html}} for model training.

\begin{table}[h]
\centering
\caption{Hyperparameters for Continued Pre-training.}
\label{tab:cpt_hyperparams}
\resizebox{.4\linewidth}{!}{%
\begin{tabular}{ll}
\toprule
\textbf{Hyperparameter} & \textbf{Value} \\
\midrule
Optimizer & AdamW \\
Learning Rate & 4e-4 \\
Min Learning Rate & 4e-5 \\
Global Batch Size & 1024 \\
Sequence Length & 4096 \\
Training Iterations & 50,000 \\
Warmup Iterations & 100 \\
Auxiliary Loss Coeff ($\lambda$) & 0.01 \\
\bottomrule
\end{tabular}%
}
\end{table}

\subsection{Supervised Fine-Tuning}
The SFT process consists of two stages with identical hyperparameters. Following the methodology of~\citep{muennighoff2024olmoe}, we use a fixed learning rate of 2e-5, a global batch size of 128, and do not use an auxiliary loss for 2 epochs. The training for both stages was conducted on 64 H100 GPUs.

\paragraph{Stage 1: General Conversational Tuning.}
This stage focuses on general conversational abilities, using a dataset mixture of LIMA \citep{zhou2023lima}, OpenHermes \citep{openhermes2023}, ShareGPT \citep{chen2024sharegpt4v}, and BAAI Infinity Instruct \citep{li2025infinity}.

\paragraph{Stage 2: Code and Math Tuning.}
This stage hones the model's capabilities in code and mathematics, using a dataset mixture of BAAI \citep{li2025infinity} and MetaMathQA \citep{yu2023metamath}, with a small amount of conversational data from Stage 1.

\section{Evaluation Datasets}
\label{app:eval_datasets}
We evaluate our models on a comprehensive suite of benchmarks to assess their capabilities across various reasoning and knowledge domains.

\begin{itemize}[leftmargin=*,itemsep=1pt]
    \item \textbf{MMLU}~\citep{hendrycks2020measuring} (Measuring Massive Multitask Language Understanding) is a broad benchmark designed to measure knowledge acquired during pre-training. It covers 57 subjects across STEM, humanities, social sciences, and more, making it a robust test of world knowledge and problem-solving ability.

    \item \textbf{HellaSwag}~\citep{zellers2019hellaswag} is a commonsense reasoning benchmark that challenges models to complete a sentence by choosing the most plausible ending from four options. It is designed to be difficult for models that rely on superficial statistical patterns.

    \item \textbf{ARC}~\citep{clark2018think} (AI2 Reasoning Challenge) is a question-answering dataset containing grade-school level science questions. We use both the Easy (ARC-e) and Challenge (ARC-c) sets, which are designed to be answerable with simple retrieval or multi-hop reasoning, respectively.

    \item \textbf{PIQA}~\citep{bisk2020piqa} (Physical Interaction Question Answering) is a commonsense reasoning benchmark focused on physical interactions. It presents two possible solutions to everyday situations, and the model must choose the more physically plausible one.

    \item \textbf{WinoGrande}~\citep{sakaguchi2021winogrande} is a large-scale dataset for commonsense reasoning, formulated as a Winograd Schema Challenge. It requires resolving pronouns in ambiguous sentences, which is challenging for models without a deep understanding of context.

    \item \textbf{LogiQA}~\citep{liu2020logiqa} is a dataset designed for logical reasoning. It consists of reading comprehension questions from professional logic exams, requiring the model to perform complex logical operations.

    \item \textbf{SciQ}~\citep{welbl2017crowdsourcing} is a question-answering dataset containing science exam questions from various domains, primarily focused on physics, chemistry, and biology.

    \item \textbf{GSM8K}~\citep{cobbe2021training} (Grade School Math 8K) is a dataset of high-quality, linguistically diverse grade school math word problems, designed to test multi-step mathematical reasoning.

    \item \textbf{HumanEval}~\citep{chen2021codex} is a benchmark for evaluating code generation. It consists of 164 programming problems with function signatures, docstrings, and unit tests to assess the functional correctness of the generated code.

    \item \textbf{IFEval}~\citep{zhou2023instruction} (Instruction Following Evaluation) is a benchmark for evaluating a model's ability to follow instructions. It consists of a set of prompts with explicit constraints that the model's response must adhere to.
\end{itemize}

\begin{figure}[h]
    \centering
    \includegraphics[width=0.6\linewidth]{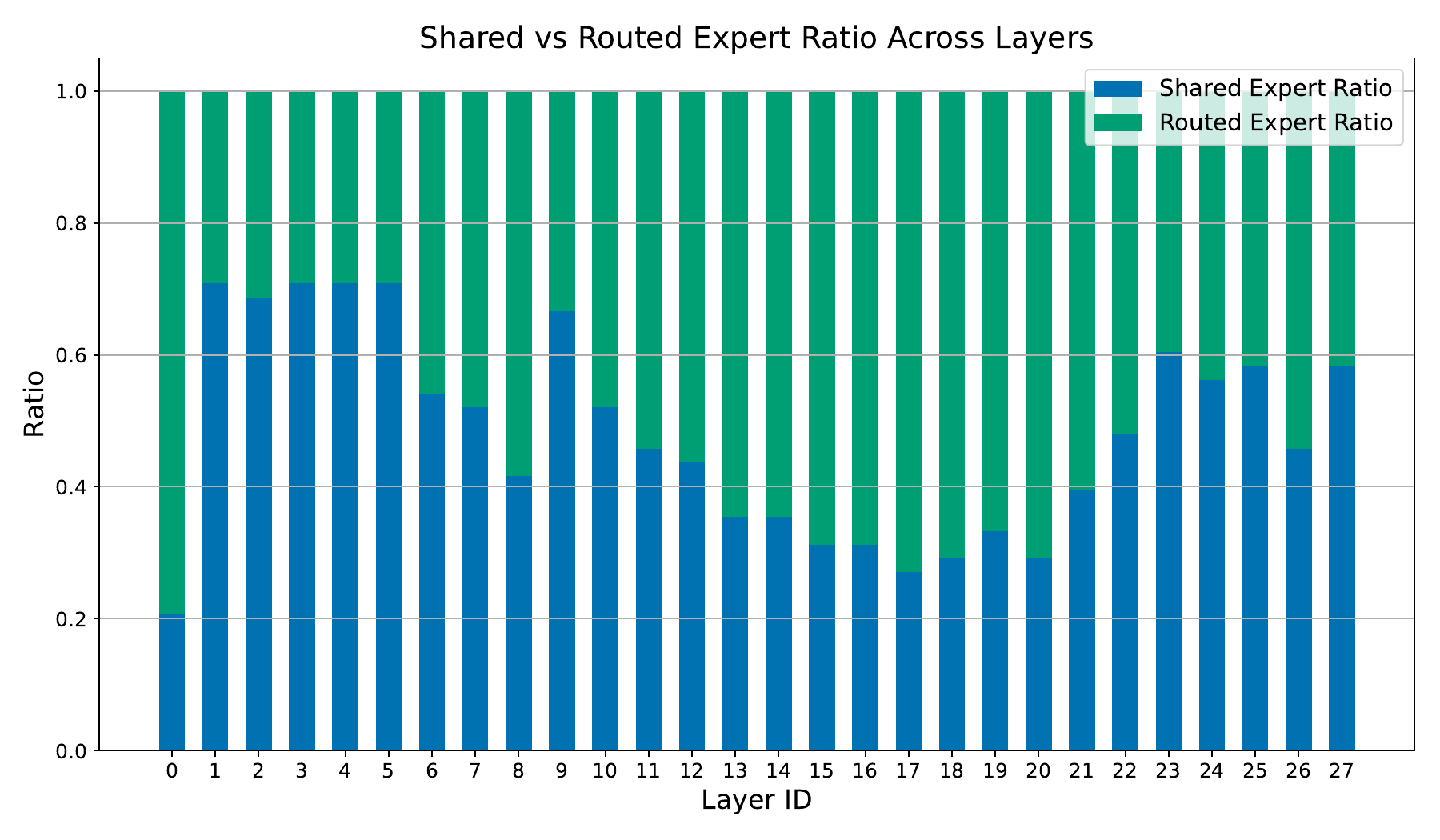}
    \caption{\textbf{Shared vs. Routed Expert Ratio Across Layers.} The figure shows the layer-wise configuration of shared and routed expert ratios for Qwen2.5-7B, as determined by ExpertWeaver.}
    \label{fig:expert_ratios}
\end{figure}

\section{Layer-wise Expert Configuration}
\label{app:layer_wise_config}
Figure~\ref{fig:expert_ratios} shows the exact configuration of ExpertWeaver at 25\% sparsity. This configuration is derived from the layer-aware expert allocation strategy in \S~\ref{sec:expert_allocation}. The resulting U-shaped distribution, with more shared experts in the initial and final layers and more routed experts in the middle layers, demonstrates ExpertWeaver's ability to automatically tailor expert composition to each layer's specific needs, contrasting with the uniform approaches used by other methods.

\begin{figure*}[h]
    \centering
\includegraphics[width=\linewidth]{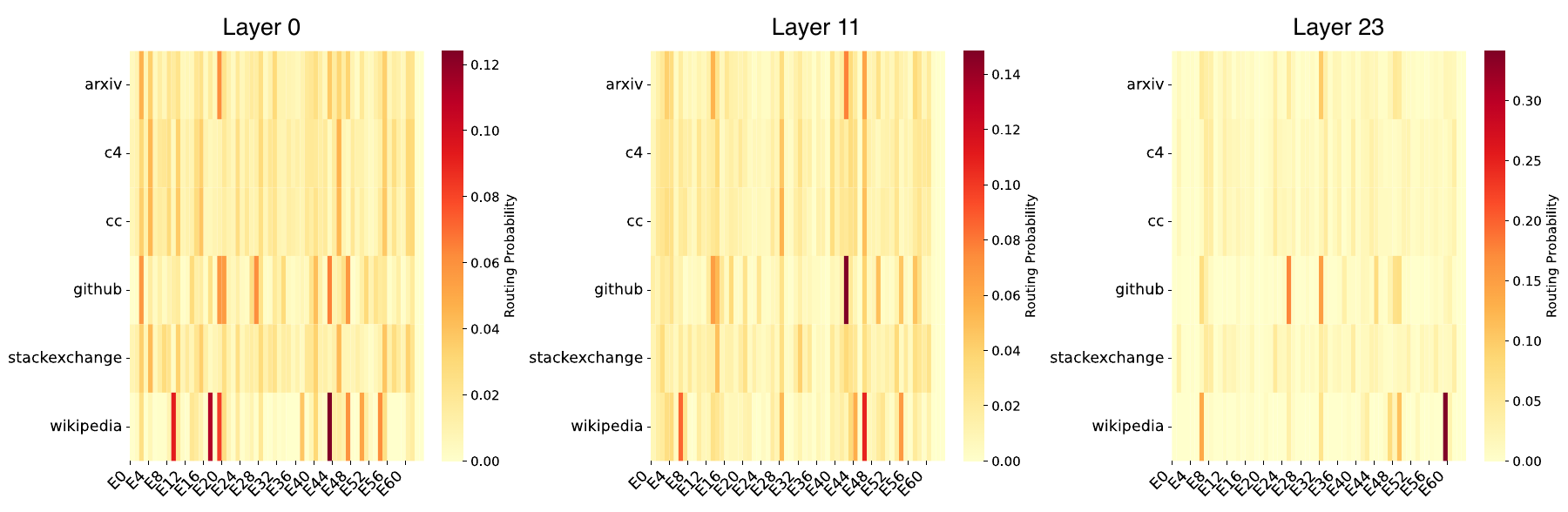}
    \caption{\textbf{Expert Specialization in the ExpertWeaver Model.}}
    \label{fig:moe_expertise}
\end{figure*}
\section{Expert Specialization}
\label{app:expert_specialization}
Figure~\ref{fig:moe_expertise} presents a detailed visualization of the expert routing patterns within our ExpertWeaver model. To create these heatmaps, we sampled 20 instances from subsets of the Red Pajama dataset (such as github and arxiv) at different model layers. The results shows that, in the shallow layers (e.g., Layer 0), tokens from various domains tend to activate a broad range of experts. Although many experts are utilized, the activation patterns between tasks are still distinguishable. The routing in deeper layers (e.g., Layer 23) becomes highly concentrated and specialized, with tokens from a specific domain consistently routed to a small and distinct set of experts. This results also reveal that shallow layers are responsible for processing common, foundational knowledge, while experts in deeper layers undergo functional differentiation to efficiently handle domain-specific information.

\end{document}